\documentclass[preprint,12pt]{elsarticle}

\usepackage{amssymb}

\usepackage{amsmath,amsfonts}
\usepackage{array}
\usepackage{textcomp}
\usepackage{url}
\usepackage{verbatim}
\usepackage{cite}
\usepackage{tabularx}

\usepackage{url}

\usepackage{graphicx}
\usepackage{multirow}
\usepackage{longtable}

\usepackage{tabularx,booktabs}
\newcolumntype{C}{>{\centering\arraybackslash}X} 
\usepackage{booktabs}

\usepackage{graphicx}
\usepackage{caption}
\usepackage{subcaption}

\usepackage{multirow}%
\usepackage{amsmath,amssymb,amsfonts}%
\usepackage{amsthm}%
\usepackage{textcomp}%
\usepackage{booktabs}%
\usepackage{subcaption}
\usepackage[table]{xcolor}
\usepackage{colortbl}
\usepackage{natbib}

\begin{document}

\begin{frontmatter}



\title{Multi-animal tracking in Transition: Comparative Insights into Established and Emerging Methods}


\author[1]{Anne Marthe Sophie Ngo Bibinbe}

\author[3]{Patrick Gagnon}

\author[1]{Jamie Ahloy-Dallaire}

\author[1]{Eric R. Paquet}

\affiliation[1]{organization={Animal Science Department, Laval University},
            city={Quebec City},
            state={QC},
            country={Canada}}

\affiliation[3]{organization={Centre de developpement du porc du Quebec (CDPQ)},
        city={Levis},
            state={QC},
            country={Canada}}

\begin{abstract}

Precision livestock farming requires advanced monitoring tools to meet the increasing management needs of the industry .Computer vision systems capable of long-term multi-animal tracking (MAT) are essential for continuous behavioral monitoring in livestock production. MAT, a specialized subset of multi-object tracking (MOT), shares many challenges with MOT, but also faces domain-specific issues including frequent animal occlusion, highly similar appearances among animals, erratic motion patterns, and a wide range of behavior types.

While some existing MAT tools are user-friendly and widely adopted, they often underperform compared to state-of-the-art MOT methods, which can result in inaccurate downstream tasks such as behavior analysis, health state estimation, and related applications. In this study, we benchmarked both MAT and MOT approaches for long-term tracking of pigs. We compared tools such as DeepLabCut and idTracker with MOT-based methods including ByteTrack, DeepSORT, cross-input consistency, and newer approaches like Track-Anything and PromptTrack.

All methods were evaluated on a 10-minute pig tracking dataset. Our results demonstrate that, overall, MOT approaches outperform traditional MAT tools, even for long-term tracking scenarios. These findings highlight the potential of recent MOT techniques to enhance the accuracy and reliability of automated livestock tracking.

\end{abstract}




\end{frontmatter}


\section{Introduction}

Animal production represents 40\% of the global agri-food sector's income. To improve the quality of animal production, the precision livestock sector, estimated at 3.2 billion dollars in 2022 and projected to reach 7 billion by 2030, is being developed to ensure effective livestock management \citep{precision_life_stock}. To do so, it will require to be able to monitor productivity, health parameters, and animal welfare in real time \citep{Schillings2021}.
One innovative approach involves using computer vision to track animals via video cameras, enabling further analyses such as behavior analysis and abnormal event detection. With recent developments in deep learning, computer vision-based tracking is promising because it is a non-invasive approach to closely monitor animal behavior over time.
As a consequence, multi-animal tracking (MAT) has garnered increasing interest in the livestock sector. Multi-animal tracking is a subclass of multi-object tracking (MOT) focused specifically on animals which involves detecting objects in each frame of a video and assigning them unique identities throughout the video.

Like MOT, MAT faces challenges such as frequent occlusions, similar animal appearances, and interactions between objects \citep{Luo_2021}. However, MAT also presents domain-specific challenges that are often more pronounced, including random and erratic movements, similar appearances, a wide range of behavioral patterns, and complex motion dynamics which might induce occlusion \citep{zhang2023animaltrack}. These factors require careful consideration when developing and applying MAT approaches.

While MOT has been extensively explored by the community, MAT remains underexplored in comparison, and there is a lack of comprehensive recent benchmarks evaluating MAT approaches in livestock particularly on long-term scenarios (eg. several minutes). On the other hand, the application of MAT in livestock has focused mainly on animal tracking tools, as seen in existing benchmarks related to animal tracking, which compare unsupervised approaches using methods such as thresholding, ellipse fitting, and filter by size for detection \citep{panadeiro2021review,wurtz2022assessment}. These approaches are often emphasized because they do not require annotations, even for the detection step, or because they offer graphical interfaces that assist non-AI specialists in performing tracking. However, these methods are less effective than methods that perform supervised detections such as the method proposed in \citep{lauer2022multi}, especially in scenarios that involve many animals.


With recent advances in MOT, several approaches have emerged that are well-suited to address these livestock-specific MAT challenges \citep{lauer2022multi, pereira2022sleap, rodriguez2018toxtrac, perez2014idTracker}. Furthermore, recent developments in foundational models have opened the door to deep learning based detection and segmentation with no need of training data, thanks to models such as SAM \citep{kirillov2023segment}, MDETR \citep{kamath2021mdetr}, and OWLv2 \citep{minderer2023scaling}. These models offer promising new directions for annotation-free tracking in livestock contexts.

Briefly, MOT is composed of two main underlying tasks : detection and identification of objects from one frame to another \citep{Luo_2021}. 

The two main tasks could further be classified as unsupervised or supervised in function of how they perform object detection and identification. The existing methods can be grouped into the following three main categories.
\begin{itemize}
    \item Unsupervised detection and unsupervised identification
    \item Supervised detection and unsupervised identification
    \item Supervised detection and supervised identification
\end{itemize}

In each category, detection and identification can be performed jointly or separately. The literature includes methods from all of these categories. However, approaches that rely on supervised identification are often impractical in the livestock context, as each new animal would require reannotation at the different stages of the life of the animal. This makes such methods difficult to apply effectively in real-world livestock settings. Nevertheless, each group of MOT approaches includes methods capable of addressing the specific challenges of MAT in livestock. 

\textbf{Contribution}
We conducted a benchmark study to compare state-of-the-art MOT and MAT approaches in livestock context across the main categories previously described, excluding methods based on supervised identification, which are less applicable in livestock contexts.

We included two widely used MAT methods: DeepLabCut \citep{lauer2022multi} and idTracker \citep{perez2014idTracker}.

We also evaluated MOT approaches with supervised detection : ByteTrack \citep{Zhang2021}, DeepSORT \citep{Wojke2018}, and Cross-Input Consistency \citep{crossinputconsistency}, as well as fully unsupervised methods such as Track-Anything \citep{yang2023track} and PromptTrack \citep{prompttrack}.

The benchmark was carried out on a 10-minute video featuring 15 active growing-finishing pigs in a pen \citep{anonymous2024an}. The video was annotated at approximately one-second intervals, with five keypoints labeled per animal (ears, nose, neck, and tail), along with individual identities. 
The goal of this benchmark is to rigorously assess the performance of leading MAT and MOT tools for long-term pig tracking. Our primary contribution is a comprehensive analysis of the strengths and limitations of each method, providing practical guidance for researchers and practitioners seeking to select and apply the most appropriate tools for livestock tracking and behavior analysis.

Our results indicate that MOT approaches (with supervised detection) outperform current MAT tools. Furthermore, among the fully unsupervised methods, Track-Anything and PromptTrack showed better performance than idTracker and produced results comparable to ByteTrack (approach with supervised detection), primarily due to the superior quality of their detection modules.

\textbf{Paper Structure }The remainder of this paper is structured as follows: Section 2 reviews related work on MAT, MOT, and existing benchmarks. Section 3 presents the experimental protocol. Section 4 reports the results, which are analyzed in Section 5 to highlight the strengths and limitations of each approach. Finally, Section 6 concludes the paper and outlines directions for future research.

\section{Background}
\subsection{Animal Tracking Benchmark}

Animal tracking benchmarks have already been proposed in the literature \citep{Zhang2022, panadeiro2021review, wurtz2022assessment,s23010532}, but unfortunately they have not fully addressed the diversity of the selected approaches (e.g., unsupervised vs. supervised detection, MAT vs. MOT). \citep{panadeiro2021review} provided a detailed comparison of MAT tools, but lacked an experimental analysis of these tools. On the other hand, \citep{wurtz2022assessment} experimentally compared four MAT tools ToxTrac  \citep{rodriguez2018toxtrac}, BioTracker \citep{monck2018biotracker}  idTracker \citep{perez2014idTracker} and \citep{branson2009high} using three sort-term one-minute videos of pigs in the livestock sector. \citep{Zhang2022} offered a broader benchmark for MOT approaches in the context of wild animal videos, encompassing a wide range of species. The dataset selection in \citep{Zhang2022} covers diverse animal species, with videos averaging 426 frames (14.2s x 30 FPS), and the longest video containing 2,268 frames (75.6s x 30 FPS). While \citep{panadeiro2021review} and \citep{wurtz2022assessment} focused only on unsupervised MAT tools, \citep{Zhang2022} explored supervised MOT approaches. In our work, we evaluate both MAT and MOT approaches on a 10-minute 25 FPS video (total of 15000 frames with 782 annotated) and also incorporate fully unsupervised MOT methods, which have emerged with recent advances in zero-shot object detection and segmentation \citep{anonymous2024an}.

 \subsection{MOT approaches}

There have been various propositions to tackle MOT in the literature, with most of the improvements being made in tracking by detection methodology. In this approach, object detection is first performed to detect the objects, and then objects are associated across frames through matching matrices. The matrices are built based on the probability that two detected objects in consecutive frames correspond to the same entity (i.e., have the same identity). The first prominent approach in this category is SORT \citep{bewley2016simple}, which uses a Kalman filter \citep{kalmanfilter} to predict the positions of objects in the next frame.  To create a correspondence matrix between consecutive frames, distances between the predicted positions and detected objects in the next frame are used as matching weights. Then it uses the Hungarian algorithm \citep{Kuhn1955} to match the detected objects on the next frame and the current frame.  The idea behind the use of the Hungarian algorithm is to provide the set of associations (assignments) with the lowest distance cost. Based on this strategy, many approaches have been developed. DeepSORT \citep{Wojke2018} improves SORT by adding appearance features as a matching criterion between objects in consecutive frames. In the two aforementioned approaches, a threshold is fixed on detection before starting the association, so detections with low confidence are not considered. However, when an object is occluded, its detection confidence score becomes low, increasing the chances of losing track of that object. To overcome this limitation, ByteTrack \citep{Zhang2021} proposed keeping low-confidence objects and matching them after high-confidence detected objects to recover missed objects. Many other approaches have improved DeepSORT, such as StrongSORT \citep{strongsort}, BoT-SORT \citep{botsort}, BoostTrack \citep{BoostTrack}, and many others, which enhance its feature extraction for appearance matching with advanced models and improve its data association methods.

Other methods have proposed joint detection and tracking approaches in a supervised manner. In this category, TransMOT \citep{10030267} proposed an approach based on graph transformers that treats tracking as a prediction problem. FairMOT \citep{zhang2021fairmot}, on the other hand, formulates tracking as a multitask learning problem with two prediction heads: one for detection and another for extracting appearance features (the re-identification (Re-ID) head). With both heads sharing the same backbone, each head contributes to improving the other during the learning stage.

To work around the lack of annotations in MOT, some have proposed self-supervised strategies for tracking individuals. In this category, some prominent works include \citep{vondrick2018tracking} and \citep{crossinputconsistency}. In \citep{vondrick2018tracking} objects are tracked by training a model to predict where the pixel colors of one frame have moved in the next frame, based on the input provided to the model. This process allows the model to detect where pixels of an object have moved and locate its new position in the next frame. In \citep{crossinputconsistency}, on the other hand, they train two Recurrent Neural Networks (RNNs) to predict object identities. The two RNNs are given different inputs, where some frames are hidden from one or the other, and they are trained to provide the same identities for objects in shared frames.

More recently, with improvements in foundational models capable of performing segmentation and detection on images, fully deep learning-based unsupervised tracking approaches have emerged. For example, Track-Anything \citep{yang2023track} leverages Segment Anyhting Model (SAM) \citep{sam} and XMem \citep{cheng2022xmem} for tracking. Similarly, PromptTrack \citep{prompttrack} uses OWLv2 \citep{minderer2023scaling} for textual prompt-based detection and ByteTrack \citep{Zhang2021} for tracking. 
\citep{cheng2023tracking} have also proposed a tracking methodology using an image-level model for segmentation, combined with a model for temporal propagation, merging the two to provide coherent tracking. With this architecture, as well as prompt-based foundational detectors, foundational models for segmentation could also be utilized. In this category, we also find \citep{nguyen2024type}, which uses text descriptions of objects and foundational models to propose generalized tracking.

\subsection{MAT approaches}
 
A lot of researchers track animals to understand their behavior and interactions inside a group. To achieve this, many MAT tools have been developed to facilitate the life of researchers. These tools are mostly user-friendly and come with many features, such as behavior analysis, labeling facilities, GUI, etc. Some prominent tools in this category include idTracker, ToxTrac, DeepLabCut, SLEAP, and others \citep{pereira2022sleap, lauer2022multi, rodriguez2018toxtrac, perez2014idTracker}. \citep{lauer2022multi} proposed a MAT methodology using the popular tool DeepLabCut. The first step in this method is pose estimation, where keypoints of the animals and the connectivity score between these points (PAF) are provided by a pre-trained model. These points are then grouped into individual animals using their affinity scores, and tracklets are created using a box or ellipse tracker. As input, DeepLabCut requires the number of animals in the scene. With this input and the tracklets, DeepLabCut is able to stitch these tracklets together to maintain the correct number of animals in the scene. This is achieved by reducing the stitching problem to an s-t cut problem, where the nodes are the tracklets and the weights between the nodes are derived from the distance in both time and space between the end of an earlier tracklet and the beginning of a later tracklet. Similarly, SLEAP \citep{pereira2022sleap} proposed a MAT tool that works similarly to DeepLabCut in that it performs keypoint detection before reconstructing individual animals using PAFs, and then tracks the animals using optical flow. On the other hand, idTracker \citep{perez2014idTracker} proposed a completely unsupervised methodology, even for detection. First, each frame is normalized to its mean to remove the background. Then, blocks with normalized intensity higher or lower than a certain threshold are assumed to represent animals. It detects animals by subtracting the background using an average model and segments them using thresholding and a filter based on the given size of the animals. The animals are then tracked using the overlap of blocks across frames, with a deep neural network learning the features of animals in case of occlusion. ToxTrac \citep{rodriguez2018toxtrac}, is a fully unsupervised tracking methodology similar to idTracker and uses adaptive thresholding to detect animals before tracking them based on distance, speed, direction, and regions of interest.

\section{Methods}
\subsection{Protocol}
For our experiments, we used the dataset from \citep{anonymous2024an}, which consists of a 10-minute video featuring 15 pigs actively interacting in a pen, annotated every second on average. There are two types of annotations in this dataset: keypoints and bounding boxes for each animal. Regarding point annotations, each pig has its two ears, neck, tail, and nose annotated when visually available. From these keypoints, a bounding box was generated as the minimum bounding rectangle. In total, 782 frames were annotated, yielding 11,730 annotations of pigs, including bounding boxes, five keypoints, and their identity \citep{anonymous2024an}.

\subsection{Selected Tracking Approaches}
For our analysis, we selected the methods listed in Table \ref{tab:methods_selected}. The purpose is to assess one SOTA (State of the art) approach in each MOT category to compare them consistently. For MAT, idTracker(version 5.2.12) and ToxTrac are SOTA approaches widely used by researchers since they are user-friendly with graphical user interfaces to assist the user. Unfortunately, we were unable to run ToxTrac on our 10-minute video due to its length. As a result, we were unable to evaluate its performance on our dataset. We also included PromptTrack (version 1.0.2) and Track-Anything (last accessed on May 2024) to our benchmark, as both are unsupervised tracking approaches leveraging OWLv2 and SAM, thus requiring no prior supervised training. The purpose of including them in the analysis is to determine if they offer a better alternative to traditional unsupervised MAT approaches, which are commonly used.

\begin{table*}
    \centering
    \begin{tabular}{c|c|c|c}
        \textbf{Method} &  \textbf{Detection} & \textbf{Identification} & \textbf{Point/boxes}\\
        idTracker & uns & uns & uns\\
        PromptTrack & uns & uns & uns\\
        Track-Anything & uns & uns & uns\\
        DeepLabCut  & sup & uns & points \\
        ByteTrack & sup & uns & boxes\\
        Cross-Input Consistency & sup & uns & boxes\\
        DeepSORT & sup & uns & boxes \\
    \end{tabular}
    \caption{MAT and MOT approaches benchmarked in this work. (sup: supervised, uns:unsupervised, boxes: using bounding boxes, points: using keypoints)}
    \label{tab:methods_selected}
\end{table*}

DeepLabCut, another widely used tool by researchers, has the particularity of detecting keypoints before reconstructing and tracking the animal. Although DeepLabCut requires annotated keypoint data for detection, we decided to compare it to other SOTA MOT approaches that also require bounding box annotations for detection to evaluate their performance in scenarios with many animals and erratic movement, such as in pig livestock. Since DeepLabCut performances depend on the number of keypoints, we assessed its performance with different numbers of keypoints : 2 keypoints (neck and tail), 3 keypoints (2 ears and tail) or 5 keypoints (nose, 2 ears, neck, and tail). DeepLabCut (version 2.3.9) was trained to detect keypoints and PAFs over 130K epochs splits of 90\% for training and 10\% for test since it is the default configuration. The detection performance are reported in Table \ref{tab:DeepLabCut_detection_performances}. DeepLabCut also requires the definition of a skeleton, for the 2 keypoint we used: (neck and tail), for 3 keypoints we used: (tail, left ear), (tail, right ear), (left ear, right ear), and for 5 keypoints we used: (nose, left ear), (nose, right ear)(nose, neck), (left ear, right ear), (neck, tail)(neck, right ear),(left ear, neck).

\begin{table}
    \centering
    \begin{tabular}{c|c|c|c}
        \textbf{Methods} &  \textbf{accuracy} &  \textbf{Recall}&  \textbf{F1 score}\\
        2 keypoints (neck + tail)  &0.92 & 0.92 & 0.92   \\
    3 keypoints (2 ears + tail)   & 0.85 & 0.85 & 0.85 \\

        5 keypoints (nose, 2 ears, neck, and tail) & 0.98 & 0.98 & 0.98  \\

    \end{tabular}
    \caption{Keypoints detection performance for the 3 tested DeepLabCut configurations (2 keypoints, 3 keypoints and 5 keypoints) by considering $IOU >0$. Here the minimal enclosing bounding box is considered to evaluate the performance.}
    \label{tab:DeepLabCut_detection_performances}
\end{table}

For the supervised tracking approaches, we selected ByteTrack (last accessed on May 2024), DeepSORT (last accessed on May 2024), and Cross-Input Consistency (last accessed on May 2024) to cover a variety of tracking approaches. All three use a pretrained detection model that was trained using bounding boxes generated with the enclosing rectangle of the 5 keypoints of each animal.  We trained a YOLOX \citep{Ge2021} model from 11700 annotations of pigs (80\% training, 20\% validation) using 500 epochs and got an mAP of 96.5\% for the training set and 95\% for the validation set.

\subsection{Evaluation metrics} 
To compare the performance of the different MOT and MAT approaches first we used IDF1  \citep{Bernardin2008}, and MOTA  \citep{Ristani2016PerformanceMA} which are evaluation metrics generally used for tracking. IDF1 is an identity-aware F1 score that evaluates how consistently a tracker maintains the correct identities over time. It differs from a standard F1 score by focusing on identity preservation, not just detection accuracy. MOTA is a global metric that penalizes missed detections, false positives, and identity switches to assess overall tracking performance.  
However, generally, when tracking is applied to animals, the objective is to perform analysis related to each animal, so it is important to also evaluate each model on its ability to give the exact identity of each animal. For this reason, the identification accuracy, recall and F1 score of each approach were also evaluated \citep{psota2022, 9217026, meena2021smart}.

\section{Results and discussion}

\begin{figure*}
    \centering
    \includegraphics[width=14cm]{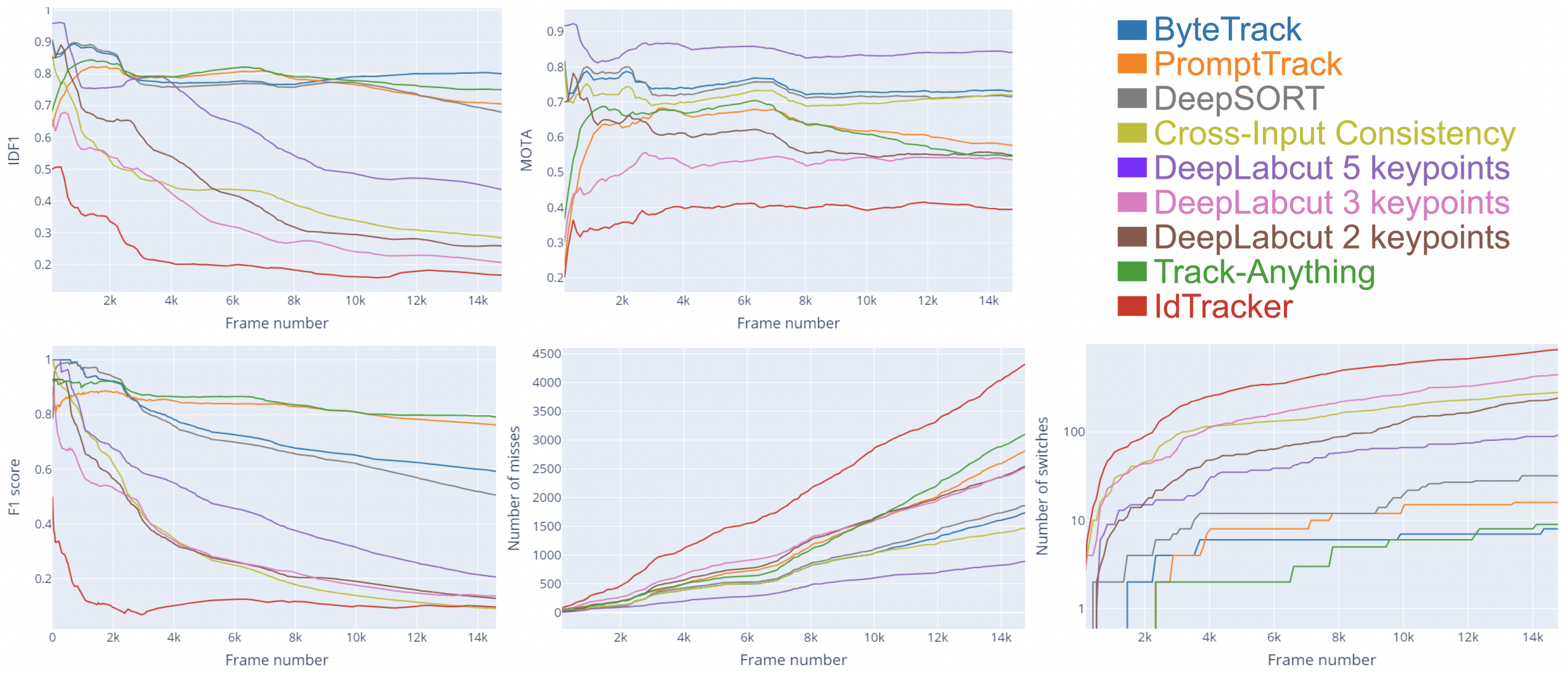}
    \caption{ Tracking performances (IDF1, MOTA, F1 score, number of misses, and number of identity switches) of the different benchmarked approaches over time in function of the number of processed frames on our long-term 10 minutes video at 25 fps.}
    \label{fig:scores}
\end{figure*}

\subsection {Comparison of supervised detection based  MAT and MOT approaches}

In this section, we compare the performance of MAT and MOT approaches that are supervised for detections. 
Among the selected methods that have a supervised detection, we benchmarked DeepLabCut, DeepSORT, ByteTrack, and Cross-Input Consistency. We noticed that the performance of DeepLabCut depend on the number of keypoints used so we tested 3 configurations: DeepLabCut with 2 keypoints (neck and tail), 3 keypoints (2 ears and tail), and 5 keypoints (2 ears, tail, neck, and nose). The F1 score, recall, accuracy, MOTA, IDF1 of those approaches tested on our 10-minute video are reported in Table \ref{tab:1}. We also estimated those performances per frame over the entire video to see how the performances evolve over time (Figure \ref{fig:scores}). As we can see from those results, the performance of most approaches decrease over time especially the F1 score, suggesting that those methods will struggle for long-term tracking. 

\begin{table}
    \centering
    \begin{tabular}{c|c|c|c|c|c}
        \textbf{Method} &  \textbf{IDF1} & \textbf{MOTA} & \textbf{F1 score} & \textbf{Recall}& \textbf{Accuracy}\\
    \midrule
ByteTrack & \cellcolor{green!79}\textbf{0.79} & \cellcolor{green!73}0.73 & \cellcolor{green!59}0.59 & \cellcolor{green!58}0.58 & \cellcolor{green!60}0.60 \\
DeepSORT & \cellcolor{green!74}0.74 & \cellcolor{green!72}0.72 & \cellcolor{green!50}0.50 & \cellcolor{green!50}0.50 & \cellcolor{green!51}0.51 \\
Cross-Input Consistency & \cellcolor{green!31}0.31 & \cellcolor{green!66}0.66 & \cellcolor{green!14}0.14 & \cellcolor{green!09}0.09 & \cellcolor{green!29}0.29 \\
DeepLabCut 2 keypoints & \cellcolor{green!32}0.32 & \cellcolor{green!56}0.56 & \cellcolor{green!13}0.13 & \cellcolor{green!13}0.13 & \cellcolor{green!13}0.13 \\
DeepLabCut 3 keypoints & \cellcolor{green!26}0.26 & \cellcolor{green!53}0.53 & \cellcolor{green!14}0.14 & \cellcolor{green!14}0.14 & \cellcolor{green!14}0.14 \\
DeepLabCut 5 keypoints & \cellcolor{green!52}0.52 & \cellcolor{green!84}\textbf{0.84} & \cellcolor{green!20}0.20 & \cellcolor{green!20}0.20 & \cellcolor{green!20}0.20 \\

\hline 

idTracker & \cellcolor{green!16}0.16 & \cellcolor{green!32}0.32 & \cellcolor{green!10}0.10 & \cellcolor{green!09}0.09 & \cellcolor{green!10}0.10 \\
PromptTrack & \cellcolor{green!66}0.66 & \cellcolor{green!48}0.48 & \cellcolor{green!76}0.76 & \cellcolor{green!74}0.74 & \cellcolor{green!79}0.79 \\
Track-Anything & \cellcolor{green!70}0.70 & \cellcolor{green!45}0.45 & \cellcolor{green!79}\textbf{0.79} & \cellcolor{green!76}\textbf{0.76} & \cellcolor{green!82}\textbf{0.82} \\
\bottomrule

    \end{tabular}
    \caption{Average tracking performance of the different benchmarked approaches over the entire 10-minute video. The color in the different cells goes from white (lower performance) to green (higher performance). The best approaches for a given metric is presented in bold.}
    \label{tab:1}
\end{table}

\begin{figure*}
    \centering
    \includegraphics[width=12cm]{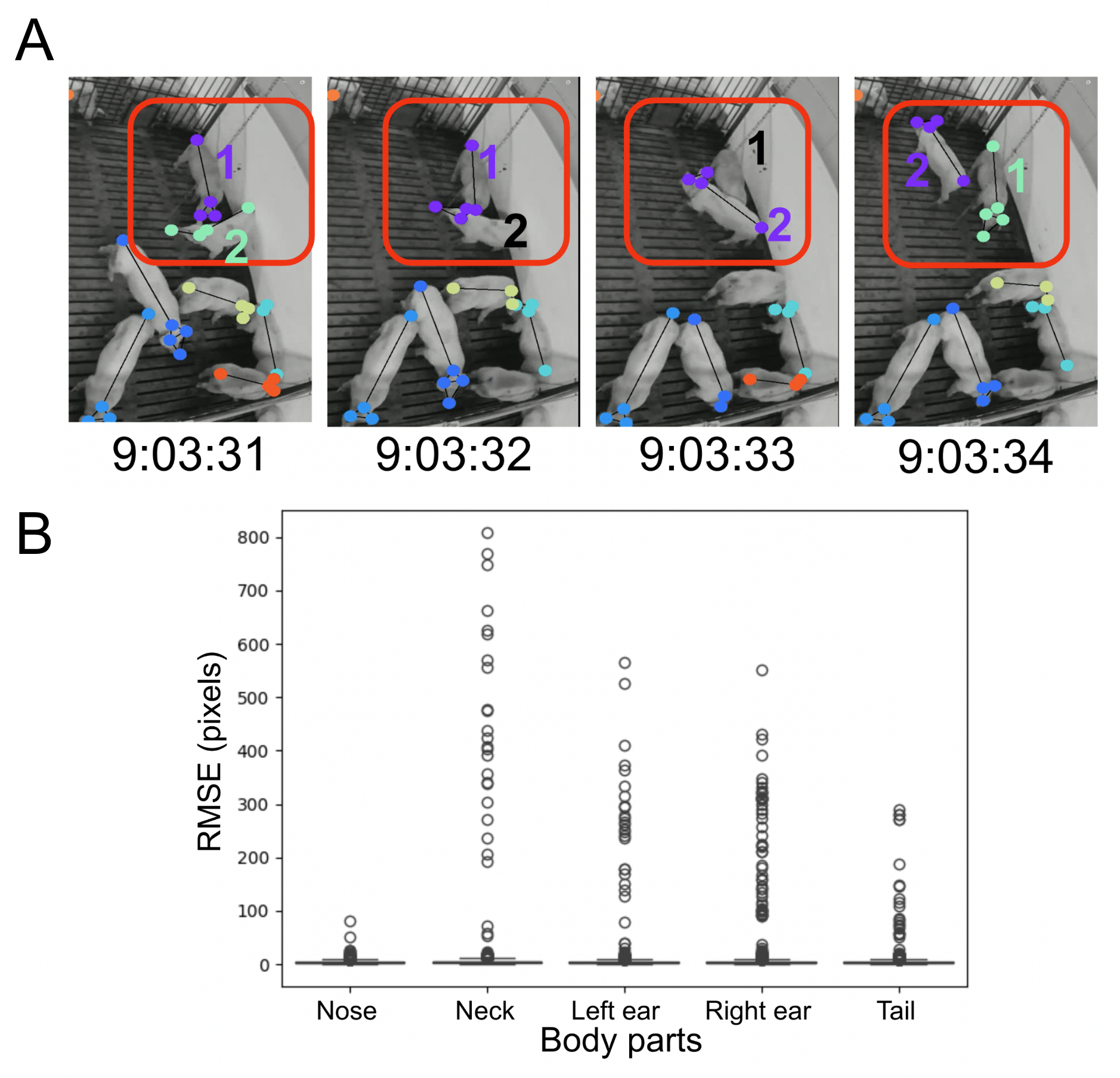}
    \caption{Illustration of DeepLabCut identity switch errors and corresponding keypoint RMSE distributions. A) Reconstruction of the skeleton of the purple pig by DeepLabCut using keypoints from another pig, which led to an identity switch. B) RMSE (or distance between the ground truth point and the detected point by DeepLabCut) in function of the different body parts. A pig is roughly 100px in our video.}
    \label{fig:DeepLabCut_combined}
\end{figure*}

\begin{table}
    \centering
    \begin{tabular}{c|c|c|c|}
        \textbf{Methods} &  \textbf{5px}  &  \textbf{10px}&  \textbf{100px}\\
        2 keypoints  & 42.35\% & 11.17\% & 0.33\%\\
      3 keypoints  & 50.44\% & 4.63\% & 0.41\%\\
        5 keypoints &  63.07\% & 7.58\% & 0.82\%\\

    \end{tabular}
    \caption{Percentage of pigs with at least one matching keypoint with a distance greater than 5, 10, or 100 pixels over the entire DeepLabCut tracking in the 10-min video.}
    \label{tab:DeepLabCut_points_taken_from_neighbor_individual}
\end{table}

The results presented in Table \ref{tab:1} show that DeepLabCut performs best when using 5 keypoints, both in overall tracking metrics and in its performance over time (Figure \ref{tab:DeepLabCut_detection_performances}). Despite this, and except for MOTA, DeepLabCut’s results with 5 keypoints remain lower than those of most supervised MOT approaches. This is notable considering that DeepLabCut requires lower-level annotations (keypoints instead of bounding boxes), which are more time-consuming to obtain before the tool can be used.

When analyzing identity switches over time, DeepLabCut shows a consistently higher number of switches compared to other approaches. Upon investigating this behavior, we found that DeepLabCut occasionally reconstructs skeletons by mistakenly incorporating keypoints from neighboring animals (Figure \ref{fig:DeepLabCut_combined}, Table \ref{tab:DeepLabCut_points_taken_from_neighbor_individual}). For instance, when a keypoint is missing from a given animal, DeepLabCut may substitute it with a keypoint from a nearby pig. Given that the average pig in our videos measures about 100 pixels in length, we identified that approximately 0.82\% of pigs had at least one keypoint located around 100 pixels away, likely originating from a different pig. From this, we estimated that on average, a pig has one misassigned keypoint every 9 frames.

While its susceptibility to keypoint misassignments results in many identity switches, DeepLabCut’s use of multiple keypoints provides robustness against missed detections of entire animals, leading to its relatively high MOTA (Figure \ref{fig:scores}).


Looking at the supervised methods in Figure~\ref{fig:scores}, Cross-Input Consistency exhibited the lowest performance overall, primarily due to a high number of identity switches. Among the approaches relying on supervised detection, ByteTrack appears to be the most promising. It maintained relatively stable IDF1 and MOTA scores over time and achieves average performance in terms of F1 score, number of misses, and identity switches. As an enhanced version of DeepSORT, ByteTrack shows similar performance, with a slight overall advantage.

Across all supervised tracking methods, we observed difficulties in maintaining consistent identities over extended periods, as indicated by declining F1 scores over time (Figure~\ref{fig:scores}). This degradation is primarily driven by an increasing number of identity switches and missed detections. Unfortunately, once an identity switch occurs, these methods are unable to recover or correct it.

\subsection
{Comparison of unsupervised detection based  MAT and MOT approaches}
We benchmarked idTracker, Track-Anything, and PromptTrack for the unsupervised detection category of trackers. None of these approaches required pre-training a model to detect animals in the videos. Their tracking performances are summarized in Table \ref{tab:1} and Figure \ref{fig:scores}.

As shown in Table \ref{tab:1}, Track-Anything and PromptTrack achieve performance comparable to ByteTrack, which yields the best results among supervised methods. In particular, Track-Anything performs well not only due to its detection capabilities but also because it uses SAM to segment each animal, which it then leverages through XMem to ensure accurate tracking.

As illustrated in Figure \ref{fig:scores}, both Track-Anything and PromptTrack outperformed idTracker, which had the lowest overall performance among all evaluated approaches.

\begin{table}
    \centering
    \begin{tabular}{c|c|c|c}
        \textbf{Methods} &  
        \textbf{F1 score} & 
        \textbf{recall} & \textbf{accuracy} \\
  
        idTracker & 0.86 & 0.77 &  0.97 \\
        PromptTrack  & 0.96 & 0.94 & 0.99  \\
        Track-Anything  & 0.95 & 0.92 & 0.99\\


    \end{tabular}
    \caption{Detection performances with $IOU>0$}
    \label{tab:detectu}
\end{table}

\begin{figure}[hbt!]
    \centering

    \begin{minipage}[b]{0.31\textwidth}
        \centering
        \includegraphics[width=\textwidth]{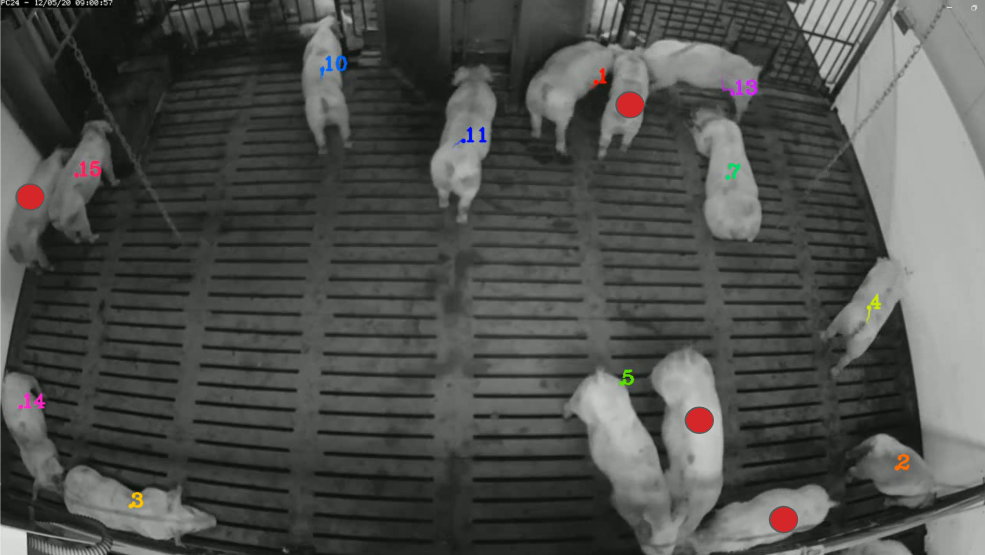}
        \label{fig:idtracker_missing}
    \end{minipage}
    \hfill
    \begin{minipage}[b]{0.31\textwidth}
        \centering
        \includegraphics[width=\textwidth]{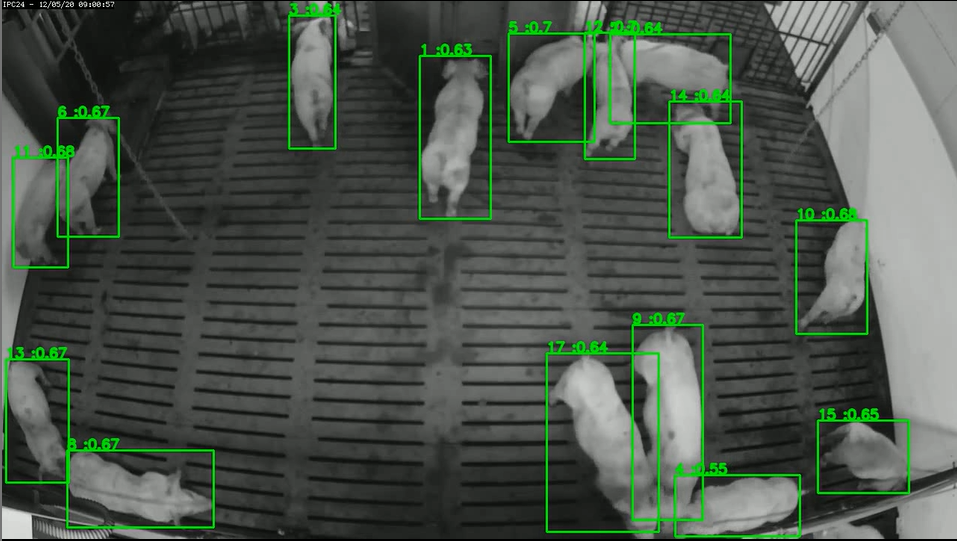}
        \label{fig:prompttrack_detected}
    \end{minipage}
    \hfill
    \begin{minipage}[b]{0.31\textwidth}
        \centering
        \includegraphics[width=\textwidth]{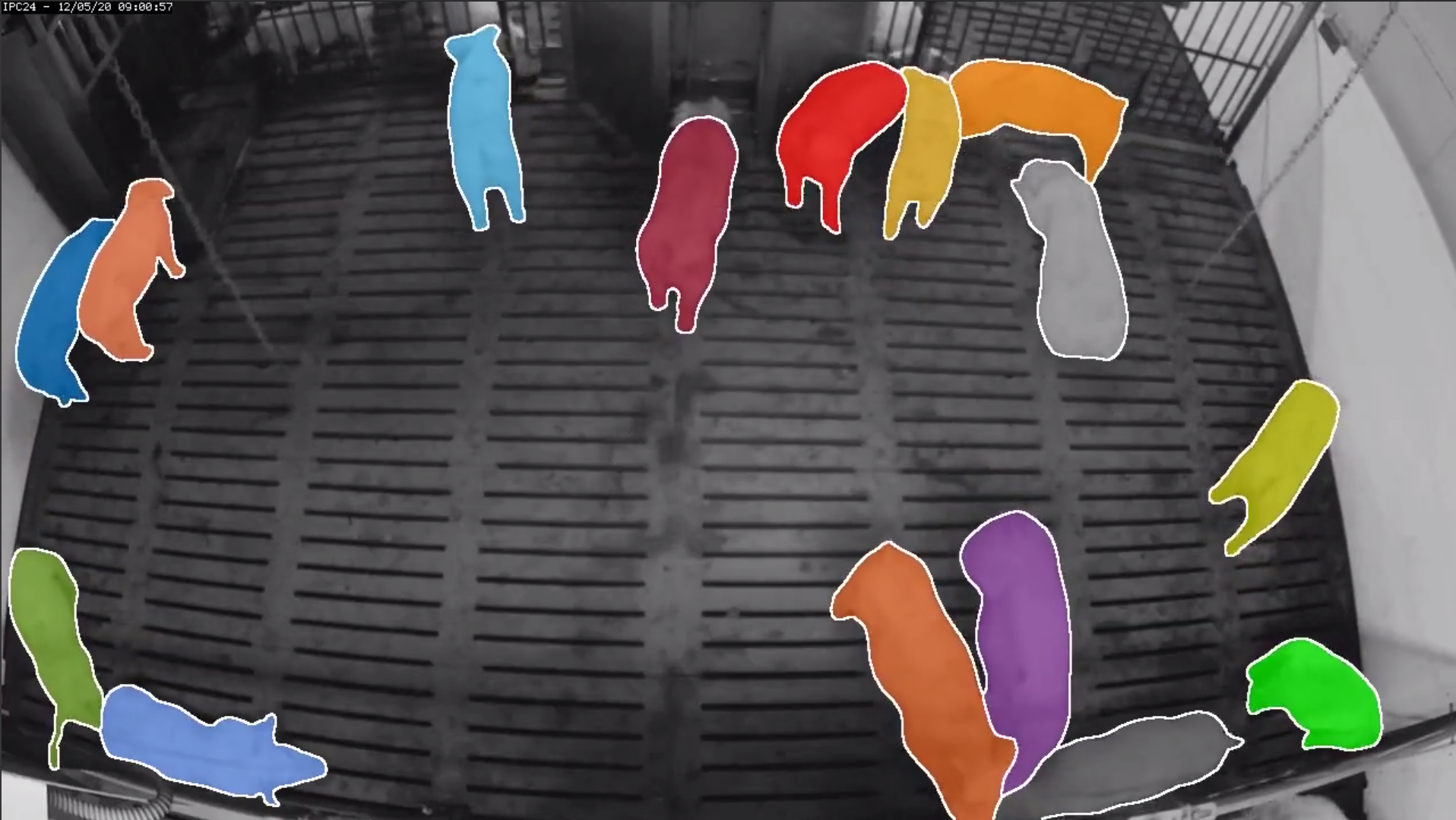}
        \label{fig:trackanything_detected}
    \end{minipage}

    \caption{ Visual comparison of detection performance between idTracker (left), promptTrack (center), and Track-Anything (right). (left) idTracker snapshot showing several undetected animals (marked with red points). (center and right) Track-Anything and PromptTrack snapshots showing all animals detected.}
    \label{fig:detection_comparison_idtracker_trackanything}
\end{figure}

We observed that the main limitation of most unsupervised approaches lies in the detection stage, which negatively impacts overall tracking performance. Detection performance was evaluated based on each method’s ability to identify ground truth boxes (recall) and its precision in each detection (precision). These results are presented in Table \ref{tab:detectu}.

As shown in Table \ref{tab:detectu}, the detection performance of unsupervised methods is generally lower than that of supervised approaches (mAP of 95\% on the validation set), particularly in terms of recall. However, Track-Anything and PromptTrack outperform idTracker in both recall and accuracy. These superior detection capabilities improved their tracking performance, as illustrated in Figure \ref{fig:detection_comparison_idtracker_trackanything}.

Currently, Track-Anything and PromptTrack do not offer standalone graphical user interfaces, which may pose a challenge for users outside the computer science field. However, both tools are relatively easy to install and use, and they benefit from active GitHub communities. In Track-Anything, the user is required to manually click on each animal in the first frame of the video. In contrast, PromptTrack only requires a text prompt, which can be especially useful for scaling to multiple videos without the need for manual annotation in each one.

\section{Conclusion}
In this study, we benchmarked several SOTA and legacy MAT and MOT approaches using the same long-term 10-minute video of pigs in a livestock setting. Our results show that MOT approaches consistently outperform MAT methods in terms of tracking performance.  They also reveal a significant challenge for all tested methods in maintaining performance over long-term tracking scenarios, as evidenced by a decline in F1 scores over time. 

Among the supervised detection-based methods, object detection was generally robust, achieving high accuracy. However, tracking performance declined over time due to frequent identity switches. Of these methods, ByteTrack provided the most balanced and reliable results. DeepLabCut, which reconstructs animal identities from detected keypoints rather than bounding boxes, struggled with detection consistency and was more prone to errors during tracking.

For unsupervised methods, detection posed a greater challenge. However, Track-Anything and PromptTrack successfully addressed this limitation by integrating SAM and OWLv2, respectively. Track-Anything, in particular, benefited from segmentation-level detection, resulting in superior tracking precision. Despite being unsupervised, both methods achieved performance comparable to ByteTrack. Additionally, they require no pre-trained detection models: Track-Anything allows users to manually select seed animals in the first frame via a simple GUI, while PromptTrack uses natural language prompts for initialization, an approach that scales well across multiple videos.

One key advantage of traditional MAT tools like idTracker is their user-friendly, standalone graphical interface, which makes them more accessible to users without a technical background. While Track-Anything and PromptTrack currently lack some of the tracking utilities found in other MAT tools, they nonetheless offer a fast, relatively accurate, and intuitive tracking solution. An important direction for future development would be the creation of standalone GUI-based versions for these new tools, enhancing their usability and adoption in non-technical research settings,and ultimately contributing to more accurate and reliable automated livestock tracking that supports improved animal welfare and productivity.

\section{Declaration of generative AI and AI-assisted technologies in the writing process}

During the preparation of this work the authors used chatGPT in order to improve language and readability. After using this service, the authors reviewed and edited the content as needed and take full responsibility for the content of the publication.

\section{Acknowledgements}
The authors want to acknowledge the financial support from MAPAQ Innov’action (IA120640 and IA120595) and all fruitful discussions with members of the Paquet lab.

\section{Data availability}
Data will be made available on request.



\bibliographystyle{elsarticle-num}
 \bibliography{elsarticle-num}





\end{document}